\documentclass[letterpaper, 10 pt, conference]{ieeeconf}  

\IEEEoverridecommandlockouts

\usepackage[space, compress, sort]{cite}

\usepackage{graphics} 
\usepackage{epsfig} 
\usepackage{mathptmx} 
\usepackage{times} 
\usepackage{amsmath} 
\usepackage{todonotes}

\usepackage{amssymb, mathtools}  
\makeatletter
\def\@opargbegintheorem#1#2#3{\trivlist
   \item[]{\bfseries #1\ #2\ (#3)} \itshape}
\makeatother

\usepackage{float}
\usepackage[breaklinks=true,bookmarks=true,colorlinks]{hyperref}
\usepackage{graphicx}
\usepackage{booktabs}   
\usepackage{multirow}   
\usepackage{xcolor}
\usepackage{multirow}  
\usepackage{adjustbox}
\usepackage{float}
\usepackage[font=small]{caption}
\usepackage{algorithm}
\usepackage{algorithmic}
\usepackage{xspace}

\newcommand{\added}[1]{{\color{black}{#1}}}
\newcommand{\arxiv}[1]{{\color{black}{#1}}}
\newcommand{\arxivedit}[1]{{\color{black}{#1}}}

\newcommand{\method}{LAE\xspace}

\title{\LARGE \bf
Latent Activation Editing: Inference-Time Refinement of Learned Policies for Safer Multirobot Navigation}

\author{Satyajeet Das$^{*}$, Darren Chiu, Zhehui Huang, Lars Lindemann$^{\dagger}$, and Gaurav S. Sukhatme$^{\dagger}$%
\thanks{$^{*}$Corresponding author. $^{\dagger}$Equal co-advisors.}%
\thanks{This work was supported in part by NSF grant IIS-2417075. S. Das, D. Chiu, Z. Huang, and G. Sukhatme are with the Department of Computer Science, University of Southern California, 
Los Angeles, CA 90089, USA. {\tt\small \{satyajee, chiudarr, zhehuihu, gaurav\}@usc.edu}. L. Lindemann is with the Automatic Control Laboratory,  ETH Zürich, 8092 Zürich, Switzerland. {\tt\small \{llindemann@ethz.ch}\}.}
}
\begin{document}

\maketitle
\thispagestyle{empty}
\pagestyle{empty}

\begin{abstract}

Reinforcement learning has enabled significant progress in complex domains such as coordinating and navigating multiple quadrotors. However, even well-trained policies remain vulnerable to collisions in obstacle-rich environments. Addressing these infrequent but critical safety failures through retraining or fine-tuning is costly and risks degrading previously learned skills. Inspired by activation steering in large language models and latent editing in computer vision, we introduce a framework for inference-time \textit{Latent Activation Editing} (\method) that refines the behavior of pre-trained policies without modifying their weights or architecture. The framework operates in two stages: (i) an online classifier monitors intermediate activations to detect states associated with undesired behaviors, and (ii) an activation editing module that selectively modifies flagged activations to shift the policy towards safer regimes. In this work, we focus on improving safety in multi-quadrotor navigation.  We hypothesize that amplifying a policy’s internal perception of risk can induce safer behaviors. We instantiate this idea through a latent collision world model trained to predict future pre-collision activations, thereby prompting earlier and more cautious avoidance responses. Extensive \added{simulations and} real-world Crazyflie experiments demonstrate that \method achieves statistically significant reduction in collisions (nearly 90\% fewer cumulative collisions compared to the unedited baseline) and substantially increases the fraction of collision-free trajectories, while preserving task completion. More broadly, our results establish \method as a lightweight paradigm, feasible on resource-constrained hardware, for post-deployment refinement of learned robot policies. Our project page with videos and code is available at \url{https://lae-robotics.github.io/}.

 \end{abstract}

\section{INTRODUCTION}
Advances in robot learning have significantly 
pushed the boundaries of autonomy, including multi robot systems, 
driven primarily by both reinforcement learning (RL) and imitation learning~\cite{ravichandar2020recent, huang2024collision, batra2022decentralized,tang2024deep}. 
Despite these successes, most learned models function as 
black boxes with limited interpretability and explainability~\cite{lipton2018mythos}. 
Enhancing specific behaviors or addressing edge cases typically demands expensive retraining or fine-tuning, involving substantial real-world data or large amounts of simulated interactions\arxiv{~\cite{rajeswaran2017learning}}. 
Furthermore, retraining carries the risk of catastrophic forgetting, where policies lose previously learned skills when adapting to new or updated task specifications~\cite{schmied2023learning, wolczyk2024fine}.
Beyond forgetting, a broader limitation arises from the asymptotic performance plateaus often observed in RL policies. Once a policy achieves strong average performance, further optimization 
usually yields only marginal gains~\cite{henderson2018deep, dulac2019challenges}. 
Closing this final remaining performance gap (e.g., from 95\% to 99.9\%) is essential for robust real-world deployment~\cite{dulac2019challenges}. 
These challenges 
motivate the need for methods that target refinement of specific policy behaviors, without incurring the costs and risks of full re-optimization.

\begin{figure}[!tbp]
    \centering
    \includegraphics[width=0.99\linewidth]{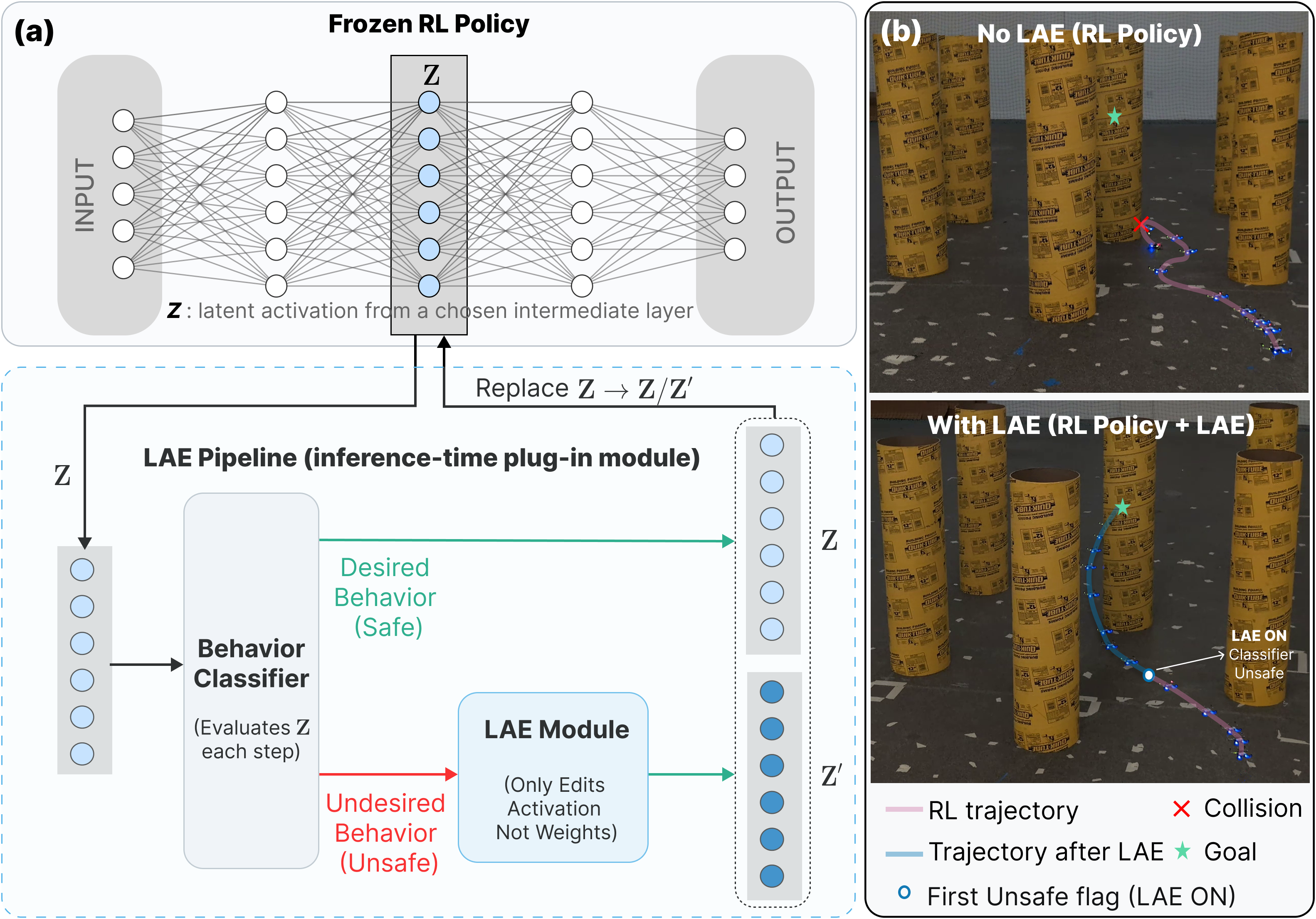}

    \caption{\textbf{(a) Conceptual overview of \method.}
    An online behavior classifier monitors the intermediate latents activation $Z$ of a frozen RL policy. Safe activations pass unchanged, while unsafe ones are replaced by edited surrogate activation $Z'$ generated by the activation editing module, without modifying policy weights. 
 \textbf{(b) Real-world quadrotor navigation illustrating \method behavior.} Without \method, the RL policy collides with an obstacle. With \method, the trajectory matches the base policy until the first unsafe flag, after which successive latent edits steer the quadrotor away from the unsafe zone, resulting in reaching the goal safely. 
    }
           
    \label{fig:hero_fig}
    \vspace{-2em}
\end{figure}

Recent advances in natural language processing and computer vision have shown that the behavior of learned models can be modified at inference time without retraining~\cite{turner2023steering,zou2023representation,templeton2024scaling, harkonen2020ganspace, meng2021sdedit}. Activation steering and representation engineering allow precise, inference-time interventions in large language models (LLMs) to guide outputs towards desired characteristics. Latent space editing in generative and diffusion models enables fine-grained control over generated content. The application of these ideas to robotics remains unexplored, primarily due to challenges associated with real-world interactions compared to text or static vision outputs.


Inspired by these advances, we propose a framework for altering the behavior of learned robot policies at inference time. 
Specifically, we focus on the problem of multi-quadrotor navigation in obstacle-rich environments, leveraging a pre-trained RL policy~\cite{huang2024collision}. While this policy achieves strong overall performance, it continues to struggle in certain edge cases and more challenging scenarios. Further retraining or architectural changes do not alleviate these failures, underscoring the need for alternative approaches~\cite{huang2024collision}.

We investigate whether targeted latent activation editing (\method) (\autoref{fig:hero_fig}) during inference can enhance safety, \added{quantified through collision avoidance}, without costly retraining or fine-tuning. 
We define \method as the process of modifying hidden activations of a network during inference, without altering its trained weights. By intervening directly in the latent space, \method temporarily adjusts the policy’s internal representations to steer behavior along desired axes, such as safety \added{(fewer collisions)} in cluttered environments.
Our key insight is that \method is a promising mechanism to reduce undesirable behaviors or enhance specific desired behaviors in pre-trained models. The specific latent dimensions chosen for editing and the underlying editing logic vary depending on the behavior to be influenced.

\method (\autoref{fig:hero_fig}) operates in two stages. First, we identify undesired states by passing the selected intermediate latent activations through an online behavior classifier. Second, we perform targeted editing of these flagged latent activations using a principled strategy. To promote safer behavior, we hypothesize that artificially amplifying the robot’s internal perception of environmental risk can trigger earlier and more cautious maneuvers, thereby improving collision avoidance. 
To realize this idea, we propose a latent collision world model (LCWM), an action-free latent world model~\cite{ha2018world,hafner2019learning,hafner2019dream} that \added{predicts} how latent activations evolve along trajectories leading to collisions, using the current activation together with a short history of past activations (Sec.~\ref{sec:lae}).
Our experiments show that among multiple \added{baseline} editing strategies, LCWM is the most effective, consistently yielding statistically superior safety performance across extensive evaluations.
Finally, we demonstrate the real-world feasibility and effectiveness of our approach through deployment on Crazyflie quadrotors, establishing \method as a practical and effective tool for enhancing the safety of pre-trained RL-based multirobot navigation policies.







Our key contributions are as follows: 
\begin{itemize}
\item  We present \method, a novel plug-in framework that steers pre-trained policies by modifying intermediate activations at inference time, enabling targeted refinement of specific behaviors. This is the first activation-space intervention demonstrated on learned robot policies.
\item We instantiate \method on the task of navigating multiple quadrotors in cluttered environments, focusing on improving the collision avoidance behavior of a pretrained RL policy.
\item We demonstrate the efficacy of \method through large-scale simulation studies and real-world quadrotor experiments, achieving statistically significant safety improvements while remaining feasible on highly resource-constrained robots.
\item Ablation studies show that effective latent editing must preserve activations relating to the robot’s own dynamics to avoid dynamically infeasible behaviors.
\end{itemize}






\section{Related Work}

\subsection{Latent Space Editing: Altering Learned Model Behavior}
Neural networks learn semantically structured internal representations that can be located and modified. Early work in natural language processing showed linear regularities and biases in word embeddings and even a “sentiment neuron” emerging under next-token training\arxiv{~\cite{mikolov2013efficient,radford2017learning}}. 
In vision, self-supervised models exhibit emergent semantics such as segmentation and depth~\cite{caron2021emerging}, motivating linear probes~\cite{alain2016understanding} and concept directions~\cite{kim2018interpretability} 
to identify editable features.
Building on this structure, latent-space editing techniques in vision successfully steer attributes by moving along interpretable directions in generative models~\cite{harkonen2020ganspace, shen2020interfacegan}
and diffusion-based guidance \arxiv{~\cite{meng2021sdedit, hertz2022prompt}}.
In LLMs, behavior can be steered via (i) prompt based methods~\cite{liu2021p},
(ii) decoding-time control~\cite{dathathri2020plugplaylanguagemodels}, (iii) weight-level knowledge editing
~\cite{meng2023locatingeditingfactualassociations}, or (iv) activation-space interventions that directly modify hidden activations during inference, such as activation addition and representation engineering~\cite{turner2023steering,zou2023representation,templeton2024scaling}.
Complementing these editing methods, representation factorization shows that activations can be decomposed into sparse, interpretable features: classical dictionary learning and sparse coding provide the foundation~\cite{elad2010sparse}
, and modern sparse autoencoders recover monosemantic features that enable causal ablations and steering
\arxiv{\cite{elhage2022toy,cunningham2023sparse,templeton2024scaling}}.
Here, we present the first application of  \added{ideas from} activation interventions to robotics. 

\subsection{Latent Representations for Safe Reinforcement Learning.} 
Recent work has explored latent representations to improve robot safety in reinforcement learning. Ls3~\cite{wilcox2022ls3} constructs latent safe sets to constrain exploration during training. LSPC~\cite{koirala2024latent} uses conditional variational autoencoders (VAEs) to enforce offline latent constraints. SLAC~\cite{hogewind2022safe} augments latent actor-critic models with a safety critic for cost-constrained training, and Latent Safety Filters~\cite{nakamura2025generalizing} apply reachability analysis in learned latent spaces to avoid unsafe actions. 
These approaches enforce safety through training-time constraints/modified architectures; we intervene directly at inference by editing activations of a frozen policy.

\subsection{Safe Multi-Quadrotor Navigation}
Safety for multi-quadrotors has been pursued 
\added{via model-based methods like}
(i) real-time planning / nonlinear model predictive control (NMPC) and (ii) control barrier functions (CBFs).
Distributed NMPC exchanges neighbors’ predicted trajectories at each control step and solves a constrained NMPC locally. This introduces communication delays and solver costs that grow with neighborhood size, horizon length, and active constraints, while assuming low-latency neighbor state information and an obstacle model~\cite{lindqvist2021scalable}. Decentralized planners such as EGO-swarm~\cite{zhou2021ego} achieve millisecond-level onboard replanning via a broadcast/chain network and local mapping, but 
would require nontrivial adaptation to interface with a learned policy. 
CBF filters solve an online \added{quadratic program} (QP) to enforce forward invariance; the seminal multi-robot CBF~\cite{wang2016safety} and graph CBF variants~\cite{zhang2025gcbf+} provide guarantees but add a runtime optimization layer, rely on accurate ego/neighbor/obstacle states, and, when paired with  RL, typically require a change of the policy architecture. 
Moreover, per-agent QPs and neighbor exchanges become a bottleneck as team size or constraints grow. 
\added{In contrast, \method is model-free and improves safety without retraining, architectural changes, or limits on the number of robots or environment clutter.}

\section{Problem Formulation}
We consider a decentralized multi-quadrotor navigation task in an obstacle-rich environment. Each robot $i$ receives local observations $O_{t, i}$ (Sec. \ref{sec:base_policy}) at time $t$ consisting of its own state $O_{t, self}$, the relative states of its nearest neighbors $O_{t,neigh}$, and a signed distance field encoding of nearby obstacles $O_{t,obst}$ (\autoref{fig:block_diagram}). A pre-trained RL policy $\pi_\phi$
\added{(specifically, the policy from~\cite{huang2024collision}) }
maps observations to rotor thrusts, \added{$a_{t,i} = \pi_\phi(O_{t,i})$}, enabling each robot to navigate to its goal while avoiding collisions. Although $\pi_\phi$ achieves strong overall performance, there is room for improvement in the collision rate. 

Our objective is to reduce collisions in $\pi_\phi$ without modifying its weights or architecture, while maintaining its goal-reaching behavior. Let $Z_t \in \mathbb{R}^d$ denote an intermediate latent representation of the policy at time $t$ where $d$ is the latent dimension.
We seek an inference-time transformation $Z_t' = \mathbf{E}_\theta(Z_t) \in \mathbb{R}^d$ that selectively replaces unsafe activations $Z_t$ with edited surrogates $Z_t'$, such that the resulting actions 
computed by the original policy with its intermediate latent $Z_t$ substituted by $Z_t'$ lead to reduced collisions while preserving goal-reaching performance.

\section{Methodology}



\begin{figure}[tbp]
\centering
\includegraphics[width=0.49\textwidth]{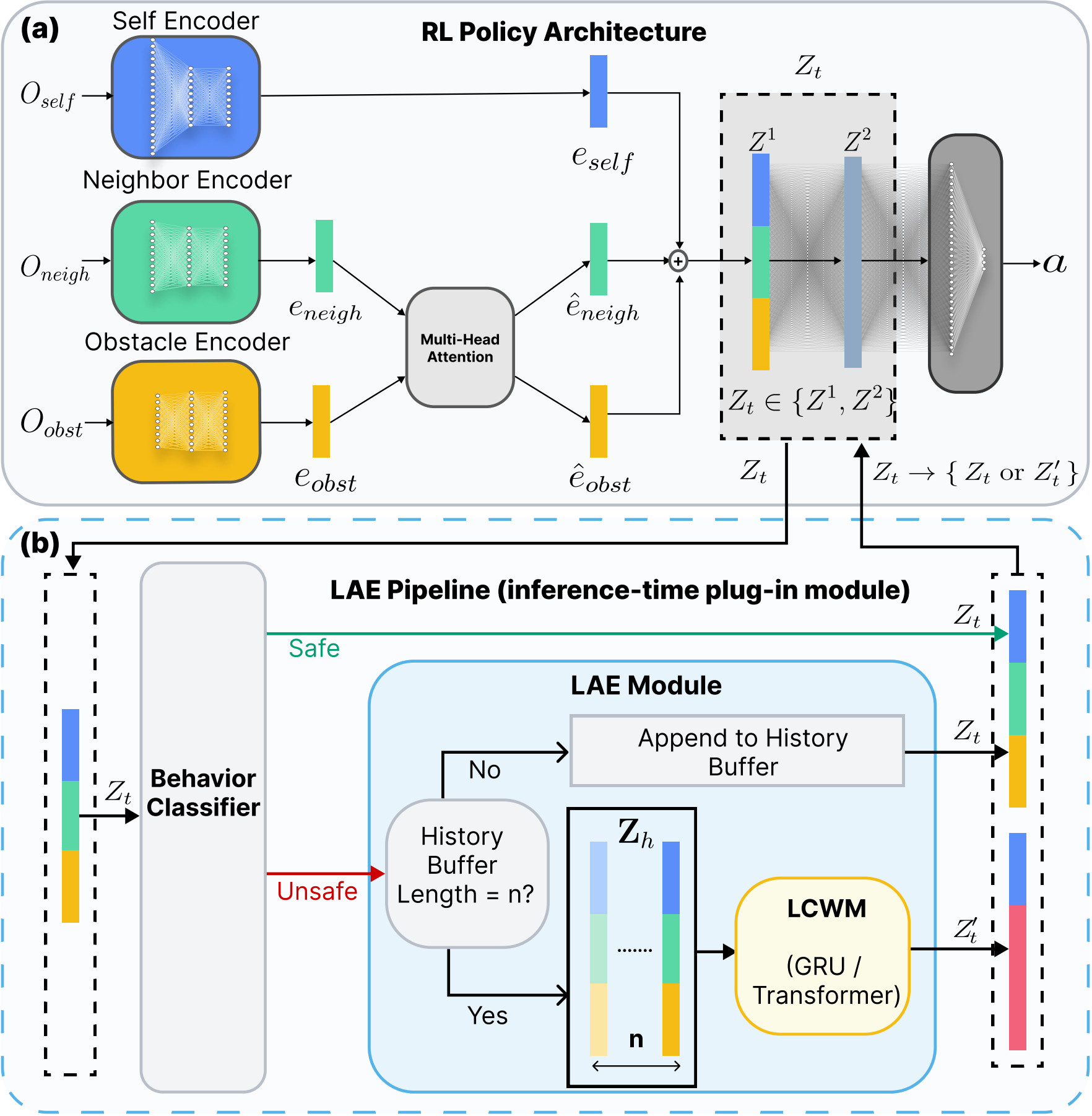}
\caption{Overview of \method integrated on a pre-trained multi-quadrotor navigation policy. \textbf{(a) RL policy architecture:}
observations are encoded and fused via multi-head attention to produce intermediate latent activations ($Z^1$, $Z^2$), which 
serve as candidates for $Z_t$. \textbf{(b) editing pipeline:} a behavior classifier evaluates 
latent activation $Z_t$ and forwards it unchanged if safe, or routes it to the editing module if unsafe. The module maintains a short history buffer and, once filled, invokes the LCWM
to generate a surrogate $Z_t'$ that replaces the unsafe latent activation $Z_t$.}
\label{fig:block_diagram}
\vspace{-1.5em}
\end{figure}

We propose an inference-time framework to modulate the behavior of a pre-trained mulit-quadrotor RL policy without altering its weights. As shown in (\autoref{fig:block_diagram}), at each timestep we extract the latent activation $Z_t$ from an intermediate encoder layer of the policy. This latent is evaluated by a behavior classifier; if it is predicted to correspond to a safe state, the policy continues unaltered. If it is predicted to be unsafe, it is replaced by an edited version $Z_t'$ generated by the latent editing module, which is then forwarded to the downstream layers of the frozen RL policy. The policy’s outputs can thus be selectively steered toward safer behaviors while retaining the original architecture and weights. The overall method consists of three components: dataset collection, behavior classifier, and latent activation editing.

\subsection{Dataset Collection}
\label{sec:dataset}
To construct the dataset, we roll out the trained policy in the QuadSwarm simulator~\cite{huang2023quadswarm}. At each policy step, we record only the intermediate latent activation \(Z_t \in \mathbb{R}^d\), together with the trajectory index \(\tau\) and time index \(t\). For each trajectory $\tau$ (\(\tau=1,\dots,K\)) of length $T$, we mark all collision time indices $t_c$ in the set \(\mathcal{C}^{(\tau)}\). Because collision times are precisely observable, we use a \emph{time-to-collision} heuristic with horizon \(H\) to obtain safety labels as
\begin{equation}
Y_{t}^{(\tau)} =
\begin{cases}
\text{unsafe} & \text{if } \exists\, t_c \in \mathcal{C}^{(\tau)} \text{ with } 0 \le t_c - t \le H,\\
\text{safe}   & \text{otherwise}.
\end{cases}
\label{eq:data1}
\end{equation}
We label a latent \emph{unsafe} if it lies within \(H\) steps of any logged collision, and \emph{safe} otherwise. This results in labeled dataset 
\begin{equation}
\mathcal{D}
= \bigcup_{\tau=1}^{K} \{(Z_{t}^{(\tau)},\,Y_{t}^{(\tau)})\}_{t=0}^{T-1},
\qquad
Z_{t}^{(\tau)} \in \mathbb{R}^{d},\;
Y_{t}^{(\tau)} \in \{\text{safe},\text{unsafe}\}
\label{eq:data2}
\end{equation}
which is used to train the behavior classifier and to construct sequence data for training the LAE (LCWM) module.

\subsection{Behavior Classifier}
The behavior classifier $\mathbf{B}_w$ is a neural network trained with supervised learning on $\mathcal{D}$ to detect unsafe activations. It implements a mapping
\begin{equation}
\mathbf{B}_w : Z_t \rightarrow \{\text{safe}, \text{unsafe}\},
\label{eq:classifier}
\end{equation}
and we denote its prediction by $\hat{Y}_t = \mathbf{B}_w(Z_t)$. We use a multilayer perceptron with batch normalization, ReLU activations, and dropout. At inference, if $\hat{Y}_t = \text{safe}$ the policy continues normally, \added{i.e., $Z_t' = Z_t$}. If $\hat{Y}_t = \text{unsafe}$, the latent activation is passed to the editing module. 
Although we focus on collision avoidance behavior, the labeling scheme is flexible; we will explore applying the same machinery to other behavioral dimensions by re-labeling $\mathcal{D}$ in future work.

\subsection{Latent Activation Editing (\method)}
\label{sec:lae}

The editing module $\mathbf{E}_\theta$ realizes our hypothesis that artificially amplifying the policy’s internal perception of risk can induce earlier and safer avoidance maneuvers. 
To put this into practice, we need a principled way to synthesize \emph{risk-amplified latent activations} consistent with the policy’s latent evolution, which in turn motivates learning a model capable of \added{predicting} 
near-future latent activations.
A \emph{world model} typically refers to a learned predictor of environment dynamics, approximating the transition function $f:(s_t,a_t)\mapsto s_{t+1}$ or, in latent form, $f:(Z_t,a_t)\mapsto Z_{t+1}$, and has been widely studied for planning and imagination in model-based RL~\cite{ha2018world,hafner2019learning,hafner2019dream}. 
Latent world models aim to capture the transition dynamics
of the environment in the latent space. However, in the multi-quadrotor navigation domain, modeling the full latent evolution proved infeasible due to partial observability, the coexistence of static and dynamic obstacles, and highly non-stationary latent dynamics.

To address these challenges, we introduce a \emph{latent collision world model (LCWM)} that focuses exclusively on latent transitions leading to collisions.
Whereas conventional world models first encode observations and condition on actions, our policy already produces latent representations, allowing us to bypass the encoder and operate directly on these activations. 
Unlike action-conditioned world models, LCWM leverages short histories of latents to capture the relevant latent evolution without explicit actions. 
Similar “action-free” latent world models have recently been proposed~\cite{schmidt2023learning}, showing that meaningful transition structure can be inferred without ground-truth actions. To this end, we train LCWM that predicts future latents along trajectories leading to collisions, using only a short history of past activations,
thereby operationalizing our safety hypothesis.

\textbf{Dataset Preparation for LCWM.}
We train LCWM using only collision-bearing trajectories (those with $|\mathcal{C}^{(\tau)}|>0$, Sec. \ref{sec:dataset}).
For each trajectory $\tau$ and collision time $t_c \in \mathcal{C}^{(\tau)}$, we consider the pre-collision window $[\,t_c-H,\,t_c\,]$.
For every index $t$ in this window, 
we form an $n$-step history buffer  of latent activations
$\mathbf{Z}_h = [Z_{t-n}, \ldots, Z_t] \in \mathbb{R}^{d \times n}$, 
where $n$ denotes the buffer length. 
To avoid predicting beyond the collision instant, we clamp the forecast index to 
$t^\star = \min(t+m,\, t_c)$, where
\added{where $m$ defines the number of steps to predict into the future.} Each $\mathbf{Z}_h$ is then paired with the target $Z_{t^\star}$, where $Z_{t+m}$ is latent 
\added{activation $m$ steps ahead of $Z_t$,}
and $Z_{t_c}$ is latent activation at the time of collision. The LCWM is trained by minimizing the objective in \autoref{eq:loss}.
\begin{equation}
\mathcal{L}(\theta) \;=\;
\sum \big\| \mathbf{E}_\theta(\mathbf{Z}_h) - Z_{t^\star} \big\|_2^2.
\label{eq:loss}
\end{equation}

\textbf{Model.} We primarily implement $\mathbf{E}_\theta$, the LCWM, as a gated recurrent unit (GRU). Given input sequence $\mathbf{Z}_h$,  the GRU evolves hidden states $h_i$ as: 
\begin{equation}
h_i = \text{GRU}(h_{i-1}, Z_i), \quad 
Z_{t}' = W h_t + b,
\label{eq:gru}
\end{equation}
where $i=t-n,\ldots,t$ and ${Z}_{t}'$ is the predicted future latent. During inference, when $\mathbf{B}_w$ flags $Z_t$ as unsafe and a buffer of length $n$ is available, the GRU predicts ${Z}_{t}'$, which replaces $Z_t$ in the forward pass. If the classifier outputs safe, the buffer is reset. 
We also implemented LCWM with a transformer-based predictor (\autoref{tab:ablations}), but the GRU variant provided slightly better empirical performance in our setting.


\textbf{Inference Loop}
The overall procedure is summarized in Algorithm~\ref{alg:inference}. At each step, the $Z_t$ is classified and either passed through or edited depending on predicted safety.

\begin{figure}[t]
\begin{algorithm}[H]
\caption{Inference-time LAE}
\label{alg:inference}
\begin{algorithmic}[1]
\STATE Predict $\hat{Y}_t \leftarrow \mathbf{B}_w(Z_t)$
\IF{$\hat{Y}_t = \text{safe}$}
    \STATE Forward latent activation, $Z_t$; reset history buffer
\ELSE
    \STATE Append $Z_t$ to history buffer
    \IF{buffer length is $n$}
        \STATE $Z'_t \leftarrow \mathbf{E}_\theta(\mathbf{Z}_h)$
        \STATE Forward new latent activation, ($Z_{t}'$)
    \ENDIF
\ENDIF
\end{algorithmic}
\end{algorithm}
\vspace{-3em}
\end{figure}
While demonstrated for collision avoidance, the framework is general: expert labeling defines the axis of intervention, the classifier acts as a representation reader, and the editor serves as a representation controller. Thus, our method can extend beyond safety to modulate other behaviors of pre-trained robot policies.



\section{Experiments}
\label{sec:experiments}

\subsection{Experimental Setup}
\label{sec:experiment_setup}

We evaluate \added{our} LAE pipeline through a combination of simulation experiments and real world deployment. In simulation, we use the QuadSwarm simulator~\cite{huang2023quadswarm} with 8 quadrotors navigating a $10 \times 10 \times 10$ m room containing static obstacles (20\% density, 0.6 m diameter). 
To ensure fair comparisons, we modified the simulator to
be deterministic,
such that identical initializations yield identical trajectories. This prevents stochastic variation and allows improvements to be attributed directly to editing.
\arxivedit{
In the default setting, major sources of stochasticity include motor \& sensor noise in QuadSwarm~\cite{huang2023quadswarm} and action sampling in Sample Factory~\cite{petrenko2020sample}. 
These components are important during training to promote policy robustness and sim-to-real transfer, but for evaluation we fix these sources to obtain reproducible comparisons (for further details see~\cite{petrenko2020sample,huang2023quadswarm}). 
}
\arxiv{Unless otherwise noted (Fig.~\ref{fig:non-deterministic}), all simulation experiments are conducted in the deterministic simulator setting.}
We construct a set of 2,600 environment configurations where the base policy collides at least once, and replay each configuration with and without LAE for evaluation. 
\added{The behavior classifier $\mathbf{B}_w$ achieves high accuracy (overall $\sim$98\%, and 99.3\% on safe trajectories), and with the additional history-buffer requirement, LAE rarely activates on already safe trajectories, where its performance is effectively identical to the baseline RL policy. Evaluation is therefore restricted to these 2600 configurations to avoid inflating results with already safe scenarios.
}
Performance is assessed using three metrics: \textbf{total collisions}, the cumulative number of total collisions across all 2600 trajectories; \textbf{zero collision trajectories}, the number of trajectories completed without any collision; and \textbf{average success rate}, the fraction of robots that reach their goals without collision, capturing task completion alongside safety. 
The remainder of this section is organized as follows. Sec.~\ref{sec:base_policy} describes the architecture and pre-training of the base RL policy. Sec.~\ref{sec:core_results} presents the core comparison between the base RL policy and our best LAE strategy - LCWM. Sec.~\ref{sec:comparative_evaluation} evaluates alternative 
\added{baseline} LAE strategies. Sec.~\ref{sec:abalation} investigates two key design questions: \emph{which latent components should be edited} and \emph{when editing should be triggered}
(i.e., how many steps before a collision). 
Sec.~\ref{sec:realworld} shows real-world deployment on Crazyflie quadrotors. 


\begin{figure}[!tbp]
    \centering
    \includegraphics[width=0.99\linewidth]{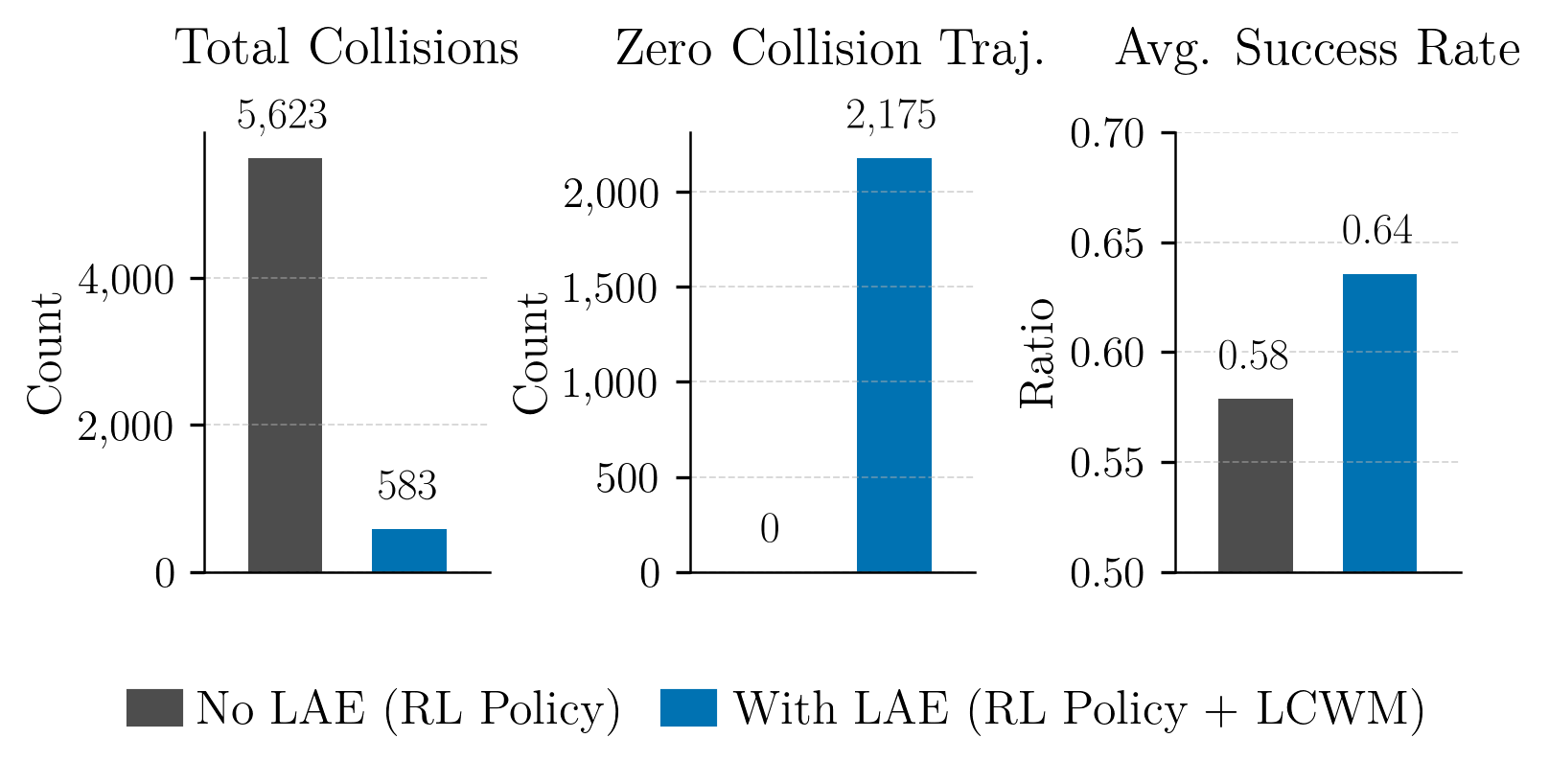}

    \caption{Quantitative comparison of the base RL policy with and without LAE on 2,600 environmental configurations. We report total collisions, zero collision trajectories, and average success rate. LAE implementation uses LCWM (GRU).}

    \label{fig:best_vs_base}
    \vspace{-1.5em}
\end{figure}

\subsection{Base RL Policy (Architecture and Pre-training)}
\label{sec:base_policy}

We adopt the end-to-end, decentralized RL policy of Huang et al.\cite{huang2024collision} as the base RL controller for the multi-quadrotor navigation task.
Each robot (\autoref{fig:block_diagram}) observes its own state and goal ($O_{t, self}$), the relative positions and velocities of its two nearest neighbors ($O_{t,neigh}$), and a compact $3 \times 3$ signed-distance field encoding of nearby obstacles ($O_{t,obst}$). 
These observations are encoded by three two-layer MLPs (self, neighbor, obstacle), with neighbor and obstacle embeddings fused through a multi-head attention module. 
The concatenated embedding produces a latent representation 
$Z^1 = [e_{\text{self}}, e_{\text{neigh}}, e_{\text{obst}}] \in \mathbb{R}^{d}$,
which is then passed through downstream MLP layers to yield another latent $Z^2 \in \mathbb{R}^{d}$. 
Finally, $Z^2$ is fed into the actor’s action parameterization head to output four normalized rotor thrusts. 
In our experiments, we instantiate $Z_t$ as either $Z^1$ or $Z^2$, i.e., $Z_t \in \{Z^1, Z^2\}$, which serve as candidate editing points for LAE (see Sec.~\ref{sec:which_latent}). The frozen RL policy used in our experiments has latent dimension $d=30$.

Training uses the asynchronous version of decentralized, independent PPO (IPPO) implemented in Sample Factory\cite{petrenko2020sample}, in a randomized 10 m × 10 m × 10 m simulated room with static obstacles and goal curricula in the QuadSwarm simulator\cite{huang2023quadswarm}. The reward encourages goal progress and penalizes robot–robot and robot–obstacle collisions and near-misses together with control regularization. A key ingredient is a collision-focused replay curriculum that buffers 1.5 s pre-impact windows at elevated sampling rates while capping long failure episodes, which reduces collision rates versus baseline variants. The learned policy scaled in simulation to 32 robots at obstacle densities up to 80 percent and transfers zero-shot for real-world deployment with Crazyflie 2.1. For more details on architecture, training setup and base-policy performance refer to Huang et al.\cite{huang2024collision}.


\begin{table*}[t]
    \centering
    \begin{tabular}{l l c c c}
        \toprule
        \textbf{Method} & \textbf{Model} & 
        \textbf{Total Collisions} $\downarrow$ & 
        \textbf{Zero Collision Traj.} $\uparrow$ & 
        \textbf{Avg. Success Rate} $\uparrow$ \\
        \midrule
        Base RL Policy (No Editing) & -- & 5,623 & 0 & 0.58 \\
        \midrule
        KD-Tree Retrieval & -- & 2,678 & 1,307 & 0.61 \\
        \midrule
        Sparse Autoencoders (SAE) & -- & 2,896 & 1,320 & 0.62 \\
        \midrule
        \multirow{3}{*}{Encoder–Decoder Projections} 
            & UMAP         & 3,766  & 491   & 0.59 \\
            & Barlow-Twins & 3,355  & 919   & 0.61 \\
            & AE           & 37,749 & 215   & 0.59 \\
        \midrule
        \multirow{2}{*}{Latent Collision World Model (LCWM)} 
            & Transformer  & 612    & 2,062 & 0.63 \\
            & GRU          & \textbf{583} & \textbf{2,175} & \textbf{0.64} \\
        \bottomrule
    \end{tabular}
 \caption{Comparison of alternative editing strategies on 2600 diverse environment configurations. 
   LCWM (GRU) achieves fewest collisions and the most zero collision trajectories while maintaining success rate. 
Other approaches offer partial gains but are less effective.}
    \label{tab:ablations}
    \vspace{-4mm}

\end{table*}
\subsection{Core Results}
\label{sec:core_results}

\autoref{fig:best_vs_base} presents the quantitative comparison between the base RL policy and 
LCWM (GRU), our best-performing instantiation of the LAE. 
The base policy incurs 5,623 collisions across the set of 2600 configurations, with no zero collision trajectories and an average success rate of 0.58. With LCWM, collisions reduce to 583 ($-89.6\%$), 2,175 trajectories ($82.7\%$) are collision-free 
, and the average success rate increases to 0.64
($+10.3\%$ relative). 
\added{This reduction in collisions is statistically significant: a paired t-test over 2,600 configurations shows a mean per-run reduction of 1.94 (95\% CI [1.86, 2.01]), $p < 10^{-300}$, Cohen's $d=1.0$.}
These results demonstrate that inference-time LAE substantially improves safety while preserving task performance. 
\added{In our experiments, we found $n=3$ to suffice, although it is task dependent. The specific choices of $Z_t$, horizon $H$, \arxiv{and prediction horizon $m$} follow the best-performing configurations identified in the ablations (Sec.~\ref{sec:abalation}).}
\begin{figure}[!tbp]
    \centering
    \includegraphics[width=0.99\linewidth]{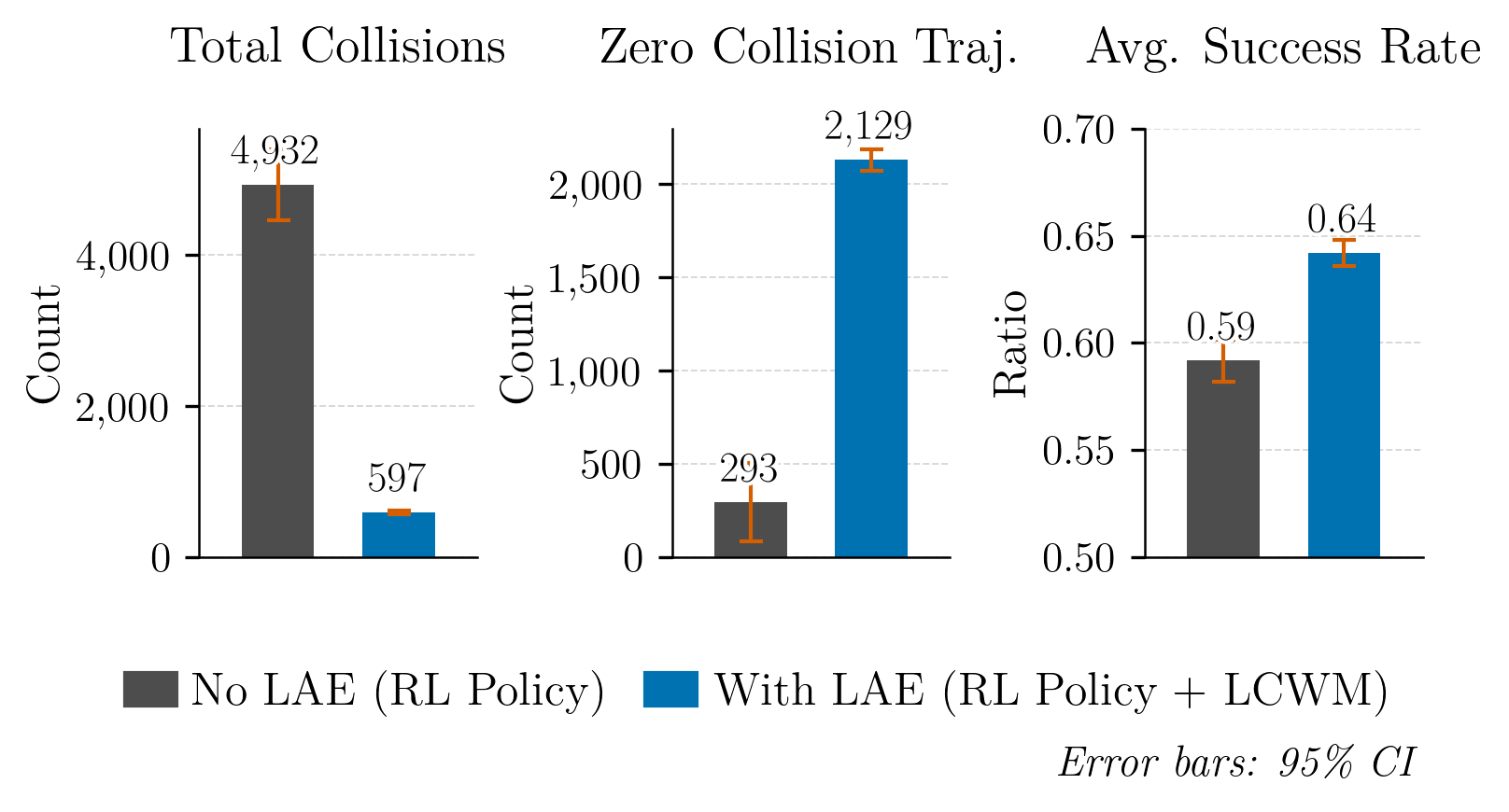}
   \caption{\arxiv{Quantitative comparison of the base RL policy with and without LAE in the non-deterministic simulator setting. We evaluate on the same 2,600 configurations with identical start–goal and obstacle locations as in the deterministic setting, but report averages over 10 stochastic runs. Error bars denote 95\% confidence intervals across runs. LAE implementation uses LCWM (GRU).}}

    \label{fig:non-deterministic}
    \vspace{-1.5em}
\end{figure}

\arxiv{While all primary evaluations in this paper are conducted in a deterministic simulator setting to ensure fair comparisons (Sec.~\ref{sec:experiment_setup}), we also performed a complementary test in the non-deterministic \arxivedit{(stochastic)} setting. This evaluation confirms that the observed improvements persist, with \autoref{fig:non-deterministic} showing similar trends: LAE substantially reduces total collisions and increases zero collision trajectories and average success rate, with error bars (95\% CI) indicating consistent performance across 10 independent runs.}

\begin{figure}[t]
    \centering
    \includegraphics[width=0.99\linewidth]{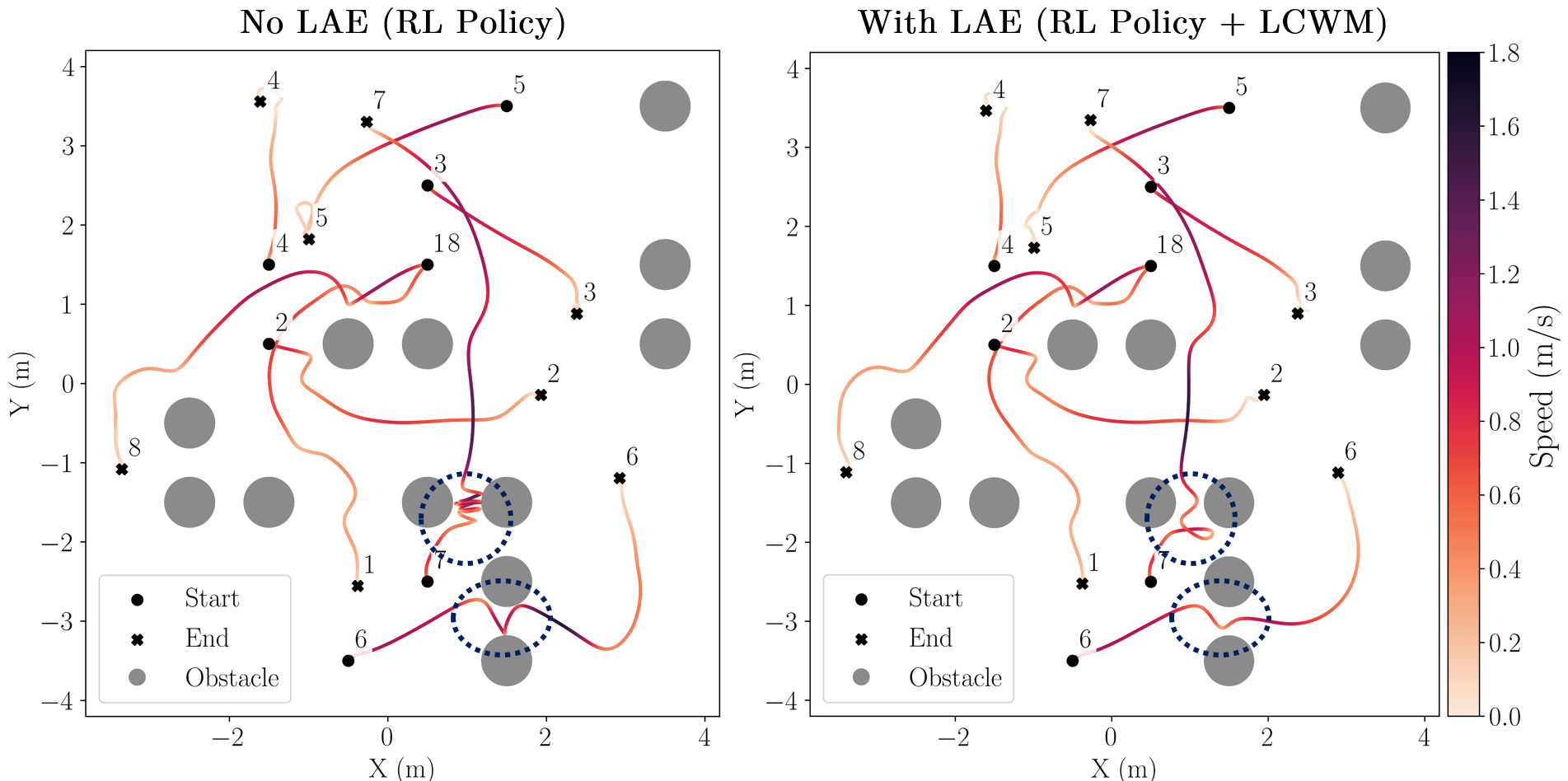}
    \caption{Representative trajectory comparison. \textbf{Left}: base RL policy (no LAE), which collides with obstacles. \textbf{Right}: RL policy with LAE, which avoids collisions while remaining identical to the base policy in safe regions and still reaching the goals. Trajectories are coloured by speed. Numbers denote quadrotor index.
    }

    \label{fig:qual_fig}
    \vspace{-1em}
\end{figure}

To complement these aggregate statistics, \autoref{fig:qual_fig} shows representative trajectory comparisons. Because editing is only triggered when unsafe states are detected, trajectories remain identical to the base policy whenever the robots operate in safe regions. When a collision is imminent, LCWM intervenes to adjust the latent state, allowing the robots to avoid the obstacle. Importantly, the agents still reach their goals, showing that the safety improvements do not come at the expense of goal-reaching behaviour.


\subsection{Comparative Evaluation}
\label{sec:comparative_evaluation}
We next compare LCWM against a range of alternative \added{baseline} LAE strategies. For each baseline, we conducted reasonable hyperparameter sweeps and report the best-performing setting we identified. While we do not claim these are globally optimal, weaker configurations performed substantially worse, and we found no settings that altered the qualitative trends reported here.
The complete results are summarized in \autoref{tab:ablations}, and we 
briefly discuss the motivation, design, and outcomes of each method.

\subsubsection{KD-Tree Retrieval}
As a non-parametric test of our LAE hypothesis for safer behavior, we built KD-trees over latent activations indexed by time-to-collision and used them to replace unsafe latents activations with their nearest neighbors drawn from trajectories leading up to a crash. 
This serves as a dataset-driven approximation of LCWM, rather than predicting future unsafe latents with a model, it simply recalls them from a dictionary of previously observed unsafe progressions.
KD-tree retrieval reduces collisions compared to the base policy (2,678 vs.\ 5,623), showing that even simple memorization of unsafe evolution provides some benefit. However, its reliance on stored examples prevents it from generalizing to unseen states, leaving it well behind the learned LCWM (583 total collisions).

\subsubsection{Sparse Autoencoders (SAE)}
\added{SAE have become popular in AI safety~\cite{cunningham2023sparse,templeton2024scaling}, where they decompose activations into interpretable units that can be selectively modulated. Following this idea, we trained SAEs on unsafe latents to identify neurons correlated with unsafe behavior and applied standard steering interventions on these units~\cite{cunningham2023sparse,templeton2024scaling}.}
While this improved safety relative to the base policy (2,896 vs.\ 5,623 collisions), it remained way less effective than LCWM. 
A key limitation is that in robotics, unsafe behaviour is distributed across entangled latent dimensions,
and may not be cleanly separable into individual sparse units
 as sometimes observed in LLMs or vision models. 
As a result, SAE performance was highly sensitive to hyperparameters such as
such as dictionary size, sparsity penalty ($\lambda$), and the choice of units targeted for editing. This fragility manifested in large performance variance, with some configurations yielding modest gains while others produced unstable or infeasible behaviour. One likely contributor is that some of the units selected for steering also encode aspects of robot's own dynamics; modifying them can disrupt the underlying stability of the controller and lead to catastrophic behaviour (Sec.~\ref{sec:which_latent}). Overall, SAEs demanded extensive tuning yet consistently underperformed the LCWM approach, highlighting their limited suitability.
\subsubsection{Encoder–Decoder Projections}
We next explored whether projecting latents into a lower-dimensional representation could act as a form of editing by implicitly mapping unsafe states closer to the manifold of safe activations. We tested parametric UMAP (structure-preserving)\added{\cite{sainburg2021parametric}},  Barlow Twins (self-supervised decorrelation)\added{\cite{zbontar2021barlow}}, and a standard autoencoder. UMAP (3,766 collisions) and Barlow Twins (3,355 collisions) provided modest improvements, but the autoencoder collapsed completely (37,749 collisions). Overall, these results indicate that while encoder–decoder projections reshape the representation space, compression alone is insufficient for effectively reducing unsafe behavior.

\subsubsection{Latent Collision World Model Variants}

The LCWM can be instantiated using different temporal latent predictors, but the input output structure and underlying hypothesis remain the same. We evaluate two widely used temporal architectures: GRUs, a popular and lightweight variant of recurrent neural networks (RNNs), and transformers. 
The GRU-based LCWM achieved the best overall performance (583 collisions, 2,175 zero-collision episodes), reflecting the suitability of recurrent models for short horizons under tight compute budgets. 
A Transformer-based LCWM achieved comparable safety (612 collisions) but incurred substantially higher computational cost, limiting its practicality for real-time deployment on Crazyflie hardware. 
These results highlight that while both architectures can realize the LCWM hypothesis, lightweight RNN such as GRUs provide the best balance of effectiveness and efficiency in our setting.



In summary, while alternative strategies offer partial gains, only LCWM consistently delivered strong safety improvements at a feasible computational cost.


\begin{figure}[t]
    \centering
    \includegraphics[width=0.99\linewidth]{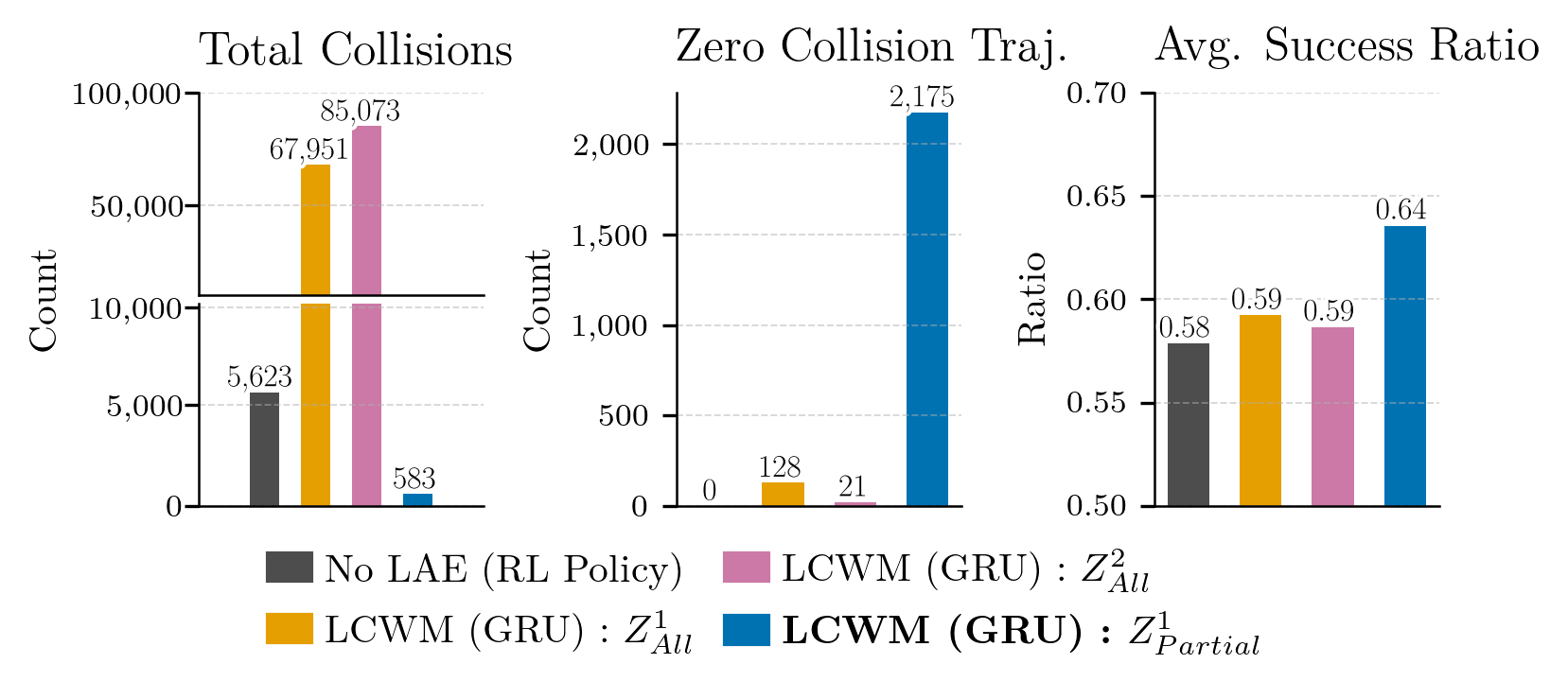}

    \caption{Choice of $Z_t$: Comparing $Z^1_{All}$, $Z^2_{All}$, and $Z^1_{Partial}$. $Z^1_{Partial}$ offers the best safety–performance trade-off.}

    \label{fig:which_latent}
    \vspace{-1em}
\end{figure}

\subsection{Ablation Studies}
\label{sec:abalation}
\subsubsection{Which Latent to Edit?}
\label{sec:which_latent}
Since LAE operates by modifying an intermediate latent activation $Z_t$, the choice of $Z_t$ is critical and a key design question.
As described in Sec.~\ref{sec:base_policy}, we consider two candidate latents: 
$Z^1$, the fused latent obtained by concatenating the self embedding with the attention-modulated neighbor and obstacle embeddings, and $Z^2$, a downstream latent before the action head.
We evaluate three editing strategies: 
(i) $Z^1_{All}$: editing the entire $Z^1$ latent activation, 
(ii) $Z^1_{Partial}$: editing only the neighbor and obstacle components of $Z^1$ while leaving the self-dynamics components untouched, and 
(iii) $Z^2_{All}$: editing the entire $Z^2$ latent activation.  
The rationale for $Z^1_{Partial}$ is that $Z^1$ retains a clean separation between self-dynamics and environment features, making selective editing possible. 
In contrast, $Z^2$ has already passed through a feedforward transformation and nonlinearity, which entangles these features and prevents a comparable split.  
As shown in \autoref{fig:which_latent}, restricting edits to $Z^1_{Partial}$ yields the best results, reducing collisions to 583, producing 2,175 zero-collision episodes, and raising success to 0.64. 
In contrast, editing the entire $Z^1$ ($Z^1_{All}$) leads to unsafe and dynamically infeasible trajectories, with performance degrading to 67,951 collisions and only 128 zero-collision episodes. 
An even stronger failure mode occurs when editing the downstream latent activation $Z^2$ ($Z^2_{All}$), the subsequent hidden layer after a feedforward transformation and nonlinearity. 
These results establish a clear guideline: effective latent editing must preserve neurons carrying self–dynamics information,
as indiscriminate modification of robot's own dynamics can produce unsafe or infeasible behavior.

\subsubsection{Effect of Editing Horizon $H$}
\label{sec:horizon}

\begin{figure}[t]
    \centering
    \includegraphics[width=0.99\linewidth]{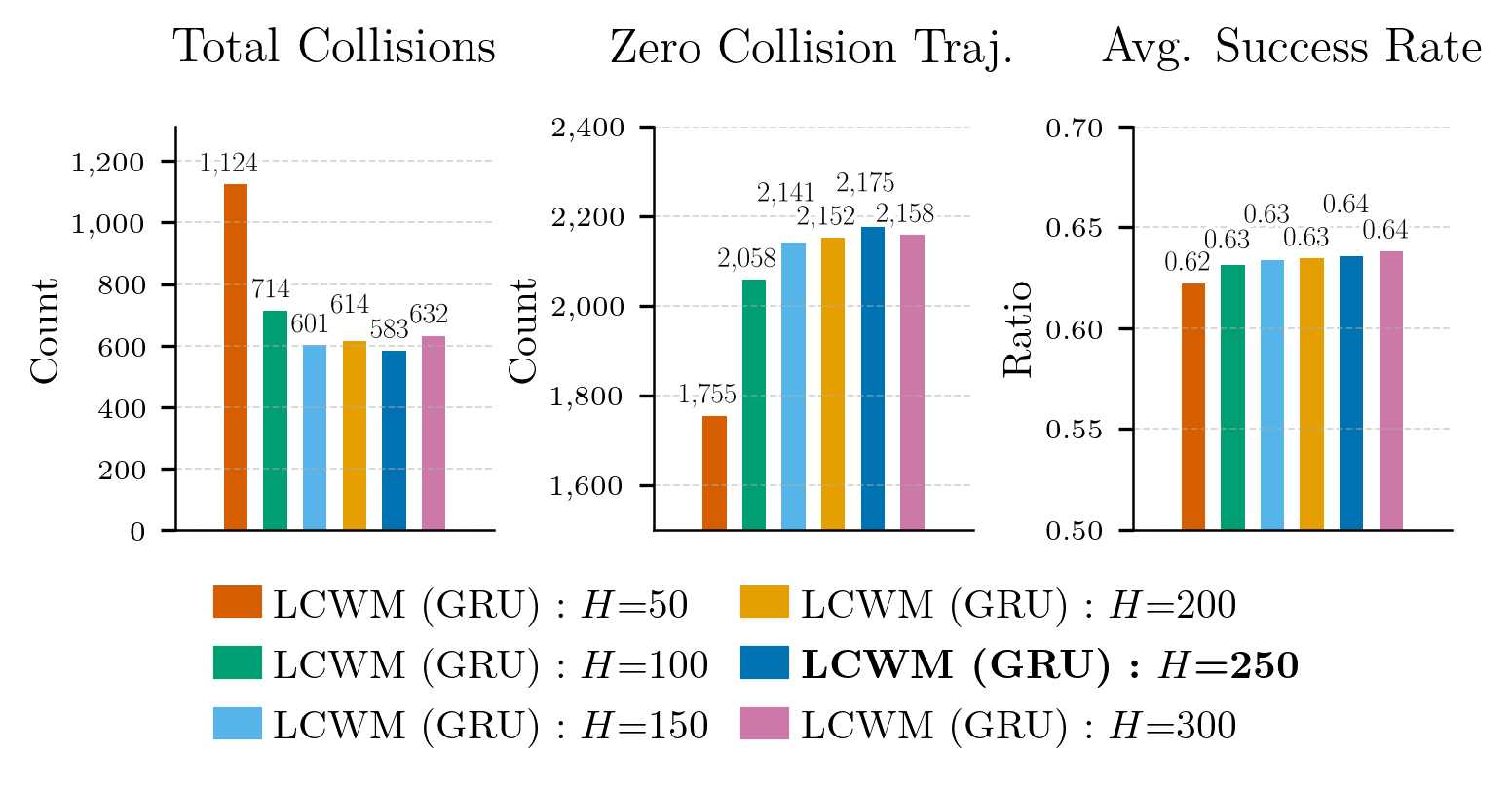}
    \caption{Effect of editing horizon $H$ for LAE with LCWM (GRU). }
    \label{fig:horizon_ablation}
    \vspace{-1.5em}
\end{figure}

The classifier horizon $H$ determines how many steps before a collision are labeled as unsafe and thus trigger editing. \autoref{fig:horizon_ablation} shows a clear trend. Short horizons ($H{=}50$) intervene too late, leaving many collisions unresolved (1,124 total). Increasing to $H{=}100$ reduces collisions substantially (714 total).
For $H\in[150,300]$, the safety and success metrics stabilize with only marginal changes, with $H{=}250$ offering the best overall trade-off (lowest collisions at comparable and highest success rate), whereas $H{=}300$ shows a slight increase in collisions. 
Concretely, total cumulative collisions drop from $1124$ at $H{=}50$ to $583$ at $H{=}250$ and total zero collision trajectories increase from $1755$ at $H{=}50$ to $2175$ at $H{=}250$  with a success rate of $0.64$. We therefore adopt $H{=}250$ as the default value. These results confirm that the timing of editing is an essential hyperparameter: even with the same model, choosing $H$ incorrectly can limit effectiveness, and the optimal value will depend on the behavior being edited.

\begin{figure}[h]
    \centering
    \includegraphics[width=0.99\linewidth]{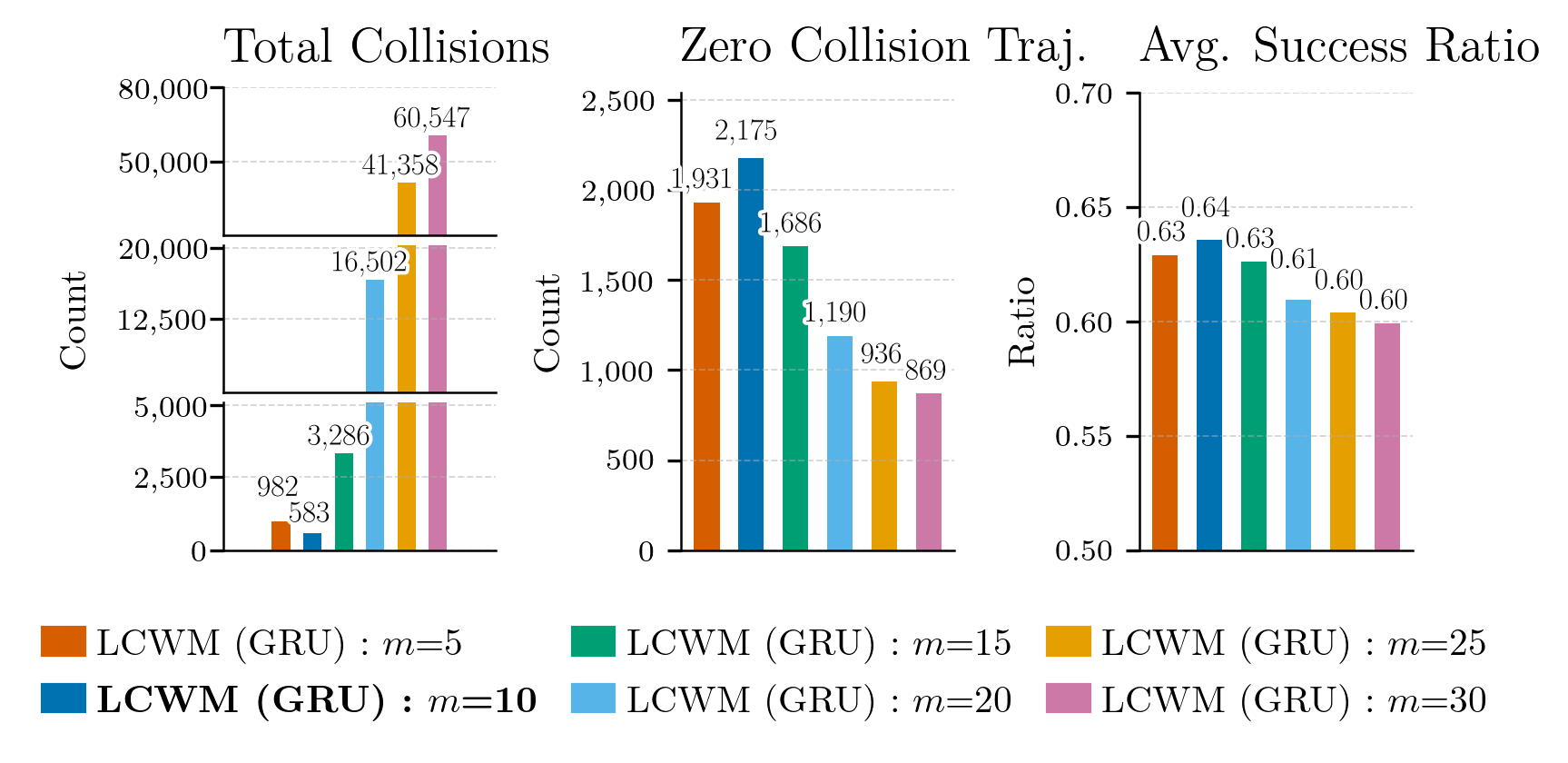}
    \caption{\arxiv{Choice of prediction horizon $m$ for LCWM.} }
    \label{fig:m_ablation}
\end{figure}

\begin{figure*}[!tbp]
    \centering
     \includegraphics[width=0.99\linewidth]{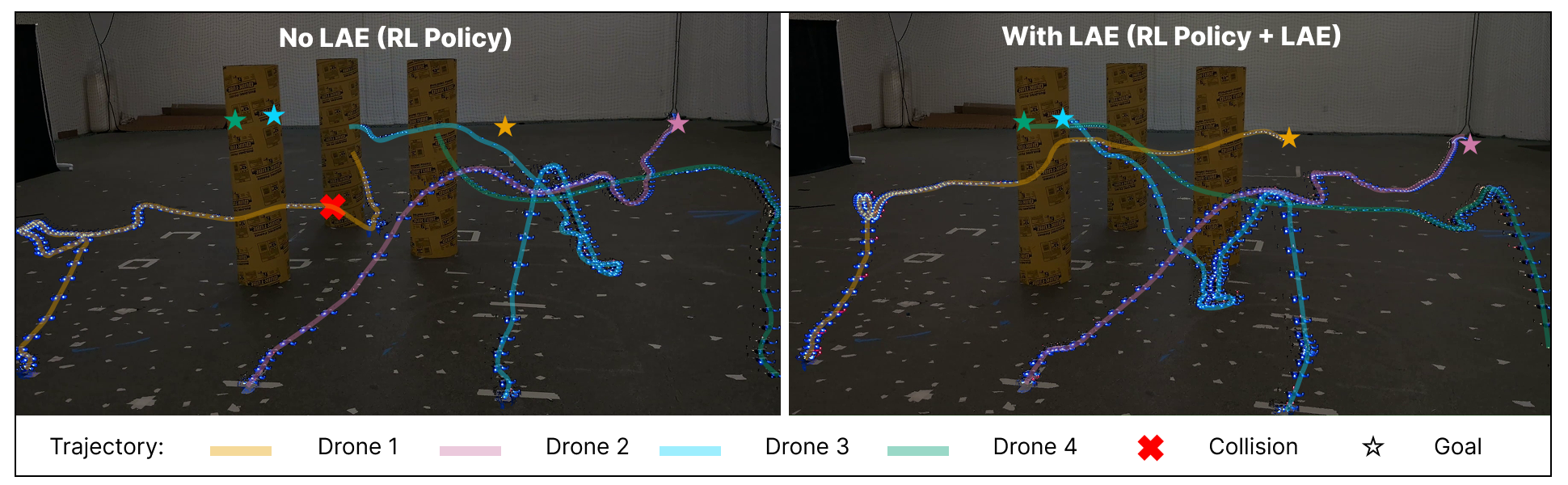}
     
\caption{\arxiv{Real-world deployment with 4 Crazyflie quadrotors navigating among cylindrical obstacles (\textbf{bilateral crossing}). 
\textbf{Left:} with the baseline RL policy, Drone~1 collides with an obstacle, leading to task failure. 
\textbf{Right:} with LAE enabled, all drones avoid collisions and reach their goals.}}



    \label{fig:realworld}
\end{figure*}
\begin{figure*}[!tbp]
    \centering
    \includegraphics[width=0.97\linewidth]{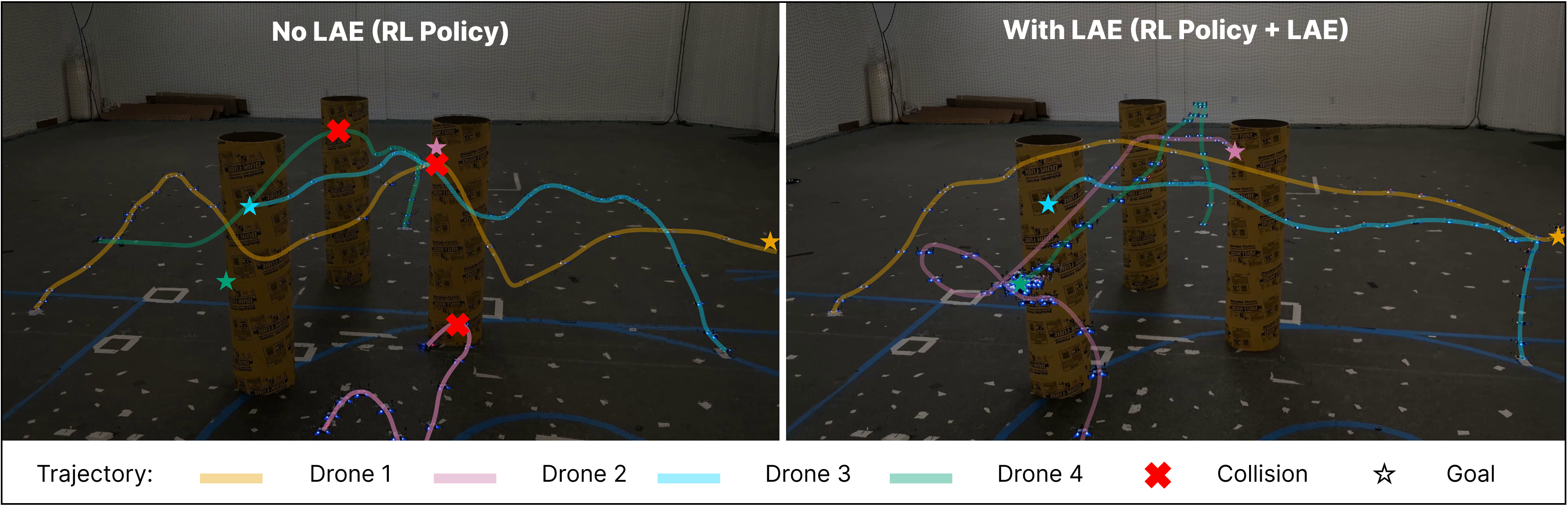}
\caption{\arxiv{Real-world deployment with 4 Crazyflie quadrotors navigating among cylindrical obstacles \textbf{(four-way crossing)}. 
\textbf{Left:} with the baseline RL policy, Drones~1, 2, and~4 collide with obstacles, leading to task failure. 
\textbf{Right:} with LAE enabled, all drones avoid collisions and reach their goals.}}

    \label{fig:realworld_1}
    \vspace{-1em}
\end{figure*}
\subsubsection{Choice of Prediction Horizon $m$}
\label{sec:pred_horizon}
\arxiv{
The LCWM operates by predicting future latent activations over a fixed prediction horizon $m$, 
which specifies how many steps into the future the model predicts. 
A sufficient lookahead is required to realize our safety hypothesis: by anticipating the near-future evolution of unsafe states, 
the LCWM can amplify collision-related activations and provide the inflated risk signal needed to trigger earlier avoidance. 
However, predicting too far ahead risks departing from the policy’s latent dynamics, 
leading to unreliable or destabilizing edits.

\autoref{fig:m_ablation} shows the effect of varying $m \in \{5,10,15,20,25,30\}$. 
Among the tested values, $m{=}10$ yields the best overall performance, 
with $583$ total collisions, $2{,}175$ zero-collision trajectories, and an average success rate of $0.64$. 
A shorter horizon such as $m{=}5$ also improves performance (1,931 zero-collision trajectories, success rate $0.63$), 
but longer horizons degrade results consistently: starting at $m{=}15$ collisions increase and success decreases, 
with the effect becoming pronounced at $m \geq 25$ (e.g., $60{,}547$ collisions at $m{=}30$). 

These results indicate that the prediction horizon $m$ is a critical hyperparameter: 
very short horizons limit foresight, while overly long horizons destabilize predictions. 
A moderate setting around $m{=}10$ provides the most favorable trade-off, 
and we adopt it as the default in for all experiments.

}

\subsection{Real-World Deployment}
\label{sec:realworld}
To validate whether LAE remains effective beyond simulation, we deploy the RL policy together with our LAE module (trained exclusively in simulation) on multiple Crazyflie~2.1 quadrotors. The LAE module consists of a compact two-layer MLP classifier with 64 hidden size ($\sim$2k parameters) and a lightweight GRU editor with hidden size 32 ($\sim$7k parameters), both re-implemented in C for real-time execution on the Crazyflie’s STM32 microcontroller. Together, these networks contain fewer than 10k parameters (under 40~kB in float32) and add less than 1~ms latency per step, making them fully compatible with the 1~kHz stabilization loop and demonstrating the feasibility of LAE on a severely resource-constrained platform.
Each quadrotor performs onboard localization through optical flow and broadcasts its estimated state to neighbors over a low-latency radio link. Obstacles are known \textit{a priori}, from which a local SDF ($2m$ range) is generated online. All computation, including state estimation, inter-quadrotor communication, LAE, and policy inference, runs fully onboard at 100~hz.
 
The quadrotors are deployed in an indoor environment with cylindrical obstacles, using identical start and goal positions across trials to ensure fair comparison. We first illustrate the LAE mechanism on a single quadrotor (\autoref{fig:hero_fig} b). The baseline RL policy collides with an obstacle, whereas with LAE the trajectory remains identical to the base policy until the first unsafe state is flagged; from that point onward, successive latent edits steer the quadrotor away from the unsafe zone while still reaching the goal safely. We then demonstrate scalability in multi-agent settings. In the 4-quadrotor bilateral crossing task (\autoref{fig:realworld}), the baseline RL policy produces collisions, whereas the LAE-augmented policy consistently steers the robots safely around obstacles while still reaching their goals. Additional experiments (\autoref{fig:realworld_1}) validate robustness in a more challenging four-way crossing setup, where quadrotors approach from all sides of the arena and must swap positions without collision. Across both scenarios, our LAE framework preserves the base policy’s goal-reaching performance while significantly reducing collisions. 
These results confirm that inference-time LAE is not only effective in large-scale simulation but also deployable as a practical, real-time safety layer for multi-quadrotor navigation on severely resource-constrained hardware.

\section{Conclusion}
We introduce \method, an inference-time framework for refining the behavior of pre-trained policies without retraining or architectural modifications. Focusing on multi-quadrotor navigation, we show that \method substantially reduces collisions by intervening directly in intermediate latent representations. We hypothesize that amplifying collision-related activations induces more cautious maneuvers and instantiate this idea through an LCWM that predicts and replaces unsafe activations. Across large-scale simulations and real-world Crazyflie deployments, \method yields statistically significant reductions in collisions while preserving task performance, establishing activation editing as a lightweight, effective, and practically feasible paradigm for post-deployment refinement of robotic control policies toward safer behavior.

While we instantiate \method for reducing collisions for multi-quadrotor navigation, the framework is not inherently limited to this setting. In future work, we plan to extend the approach to other behavioral axes, different platforms, and varied task domains.










\bibliographystyle{IEEEtran}
\bibliography{citations.bib}

\end{document}